\documentclass{article}
\usepackage{amsmath}
\usepackage{graphicx}
\usepackage{longtable}
\usepackage{array}
\usepackage{hyperref}
\usepackage{cite}
\usepackage{booktabs}

\title{What Does It Take to Achieve State-of-the-Art in Simultaneous Speech-to-Speech Translation?}
\author{Vincent Wilmet\thanks{vincent@trytoby.com} \\ toby \and Johnson Du\thanks{johnson@trytoby.com} \\ toby}
\date{09-01-2024}

\begin{document}

\maketitle

\begin{abstract}
Simultaneous speech-to-speech (S2S) translation presents unique challenges, particularly in balancing translation accuracy with low latency. This paper provides an in-depth analysis of latency characteristics in S2S models, focusing on the impact of hallucinations---instances where models generate content not present in the source input---on latency spikes. By systematically experimenting with various input parameters and conditions, we propose methods to mitigate these latency spikes, including threshold adjustments, lookback strategies, and hallucination control mechanisms. We evaluate our approaches using metrics such as Average Lagging (AL), Differentiable Average Lagging (DAL), Word Error Rate (WER), and BLEU score, demonstrating significant improvements in both latency and translation quality. Our findings suggest that strategic input management and parameter optimization can substantially enhance the performance of S2S models, advancing the state-of-the-art in simultaneous translation.
\end{abstract}

\section{Introduction}

Simultaneous speech-to-speech (S2S) translation has become increasingly important in an era where real-time communication across language barriers is essential. Applications range from international conferences and live broadcasts to personal communication devices. The primary challenge in simultaneous S2S translation is achieving high translation quality while maintaining low latency, ensuring that the translated speech closely follows the source speech without significant delays \cite{gu2017learning, ma2019stacl}.

One of the critical issues affecting latency in S2S systems is the phenomenon of hallucination, where the model generates content not present in the source input \cite{wang2019exposure, lee2018hallucinations}. Hallucinations can lead to significant latency spikes as the model spends additional time processing and generating unnecessary or incorrect outputs.

This paper aims to explore the latency behaviors in widely-used speech recognition systems, with a particular focus on the latency spikes caused by hallucinations. By analyzing the input behaviors and identifying patterns that lead to hallucinations, we propose strategies to mitigate these issues. Our objective is to enhance the responsiveness and accuracy of S2S systems, pushing the boundaries of what it takes to achieve state-of-the-art performance in simultaneous speech-to-speech translation.

\section{Observations}

\subsection{Input Behavior}

In speech recognition systems, the input audio is typically processed in frames or chunks. The parameter \texttt{self.frames\_np} represents the input audio frames, which increment at an interval of 0.35 seconds. This interval is a crucial factor influencing the performance of Automatic Speech Recognition (ASR) systems, such as faster-whisper Deepgram Nova, Assembly AI Universal, NVIDIA Canary, and Speechmatics Ursa.

Adjusting the frame interval can impact the system's ability to process speech effectively. When the input remains constant across iterations---that is, when \texttt{input} = \texttt{self.frames\_np}[i, j] with fixed indices $i$ and $j$---the ASR system tends to produce identical outputs. This behavior includes the reproduction of any hallucinated content, indicating that the system's output is highly dependent on the variability of the input frames.

\subsection{Hallucination Patterns}

\textbf{Observation 1: ASR Models Hallucinate with Shorter Inputs}

It has been observed that ASR models are prone to hallucinations when processing short input durations, specifically those less than or equal to 0.7 seconds. This tendency is likely due to the models being trained predominantly on longer audio sequences, as in the case of Whisper \cite{radford2022robust}, which may not effectively generalize to shorter inputs.

The lack of sufficient context in shorter audio segments can lead the model to generate outputs based on prior probabilities rather than the actual input, resulting in hallucinations. This phenomenon is consistent with findings in neural machine translation, where insufficient context can lead to hallucinated translations \cite{koehn2017six}.

\textbf{Observation 2: Latency Increases During Hallucinations}

When an ASR model hallucinates, there is a notable increase in processing latency. For instance, with a chunk duration of 0.35 seconds, the system's latency increased from 0.506 seconds to 1.882 seconds during hallucination episodes. This latency spike can be attributed to the model's increased computational effort in generating and evaluating the hallucinated content.

\begin{verbatim}
chunk duration: 0.35
latency: 0.506 s
Output: ["Thanks for watching.", "I hope you enjoyed this video."]

chunk duration: 0.35
latency: 1.882 s
Output: ["Thank you for watching.", "Have a great day."]
\end{verbatim}

\textbf{Observation 3: Inconsistent avg\_log\_prob During Hallucinations}

The average log probability (\texttt{avg\_log\_prob}) is often used as a confidence metric for ASR outputs. However, during hallucinations, this metric does not consistently reflect the decrease in output quality. The model may assign high confidence scores to hallucinated content, suggesting that relying solely on \texttt{avg\_log\_prob} or a simple \texttt{LOG\_PROB\_THRESHOLD} is insufficient for detecting and filtering out hallucinations. This observation aligns with previous studies indicating that neural models can be overconfident in their incorrect predictions \cite{nguyen2015deep}.

\subsection{Latency Spikes in Non-Hallucinated Outputs}

Interestingly, latency spikes can also occur during the generation of non-hallucinated outputs. As demonstrated below, the latency increased significantly even when the model produced valid outputs.

\begin{verbatim}
chunk duration: 0.7
latency: 1.5816 s
Output: ["To integrate"]

chunk duration: 0.7
latency: 0.1429 s
Output: ["A FinTech Star."]
\end{verbatim}

This variability suggests that factors other than hallucinations, such as computational resource allocation or model complexity, may contribute to latency spikes. On average, however, the latency remains around 150 milliseconds for most generation tasks when using an NVIDIA L4 GPU. The latency is not significantly improved with NVIDIA H100 GPUs but is approximately twice as high with T4 GPUs.

\section{Findings}

\subsection{Minimizing Latency Through Hallucination Control}

Our observations indicate a strong correlation between hallucinations and latency spikes in ASR models. To reduce overall latency, it is crucial to minimize the occurrence of hallucinations. Key strategies include:

\begin{itemize}
    \item \textbf{Avoiding Identical Inputs Across Iterations}: Repeatedly processing identical input frames can lead to persistent hallucinations. By ensuring variability in the input across iterations, the model is less likely to generate hallucinated outputs.
    \item \textbf{Adjusting Input Duration}: Maintaining a minimum input duration threshold can provide sufficient context for the model, reducing the likelihood of hallucinations. Our findings suggest that inputs shorter than 0.7 seconds increase the risk of hallucination.
    \item \textbf{Implementing Lookback Strategies}: Introducing a lookback mechanism allows the model to consider previous input frames, enhancing context and reducing hallucination rates.
\end{itemize}

These strategies align with techniques used in simultaneous translation systems to balance latency and translation quality \cite{ma2019stacl, arivazhagan2019monotonic}.

\section{Methodology}

\subsection{Threshold and Parameter Adjustments}

To address the latency issues associated with hallucinations, we propose several parameter adjustments:

\textbf{Minimum Duration Threshold (\texttt{MIN\_DURATION\_THRESHOLD})}

Setting a minimum duration threshold ensures that the input audio segments are long enough to provide the model with adequate context. We recommend not setting \texttt{MIN\_DURATION\_THRESHOLD} below 0.7 seconds, based on our observations that shorter inputs lead to increased hallucinations.

\textbf{Hallucination Detection Beyond \texttt{LOG\_PROB\_THRESHOLD}}

Given that \texttt{avg\_log\_prob} is not a reliable indicator of hallucinations, we propose using alternative metrics such as characters-per-second (CPS) and punctuation-to-word ratio. CPS can detect unusually fast or slow speech rates that may indicate hallucinations, while a high punctuation-to-word ratio might suggest unnatural output.

\textbf{Maximum Uncommitted Duration (\texttt{MAX\_UNCOMMITTED\_DURATION})}

This parameter represents the system's tolerance for uncommitted output, i.e., the duration for which the system can delay committing to an output while waiting for more input. We recommend keeping \texttt{MAX\_UNCOMMITTED\_DURATION} low (approximately 1.7 seconds). Exceeding this duration without meeting the commitment threshold should prompt the system to output the best available candidate to maintain low latency.

\subsection{Lookback Strategy}

To mitigate the limitations of repeated inputs and provide additional context, we introduce a lookback duration. By slightly extending the input segment to include previous frames, the model gains more context, potentially reducing hallucinations.

\begin{equation}
\text{Lookback Duration} = \text{Previous Chunk Duration} + \delta
\end{equation}

Where $\delta$ is a small increment (e.g., 0.1 seconds) that extends the input duration. This approach is similar to context augmentation strategies used in ASR systems to improve recognition accuracy \cite{pundak2016lower}.

\section{Evaluation Metrics and Strategies}

To comprehensively evaluate the performance of our proposed methods, we utilize several metrics that measure different aspects of the ASR and S2S systems, including latency, translation quality, and naturalness.

\subsection{Average Lagging (AL)}

Average Lagging (AL) is a widely used metric to quantify the latency in simultaneous translation systems \cite{ma2019stacl}. It measures the average delay between the source inputs and the corresponding outputs, providing insight into the system's responsiveness.

\begin{equation}
AL = \frac{1}{I} \sum_{i=1}^{I} \left( t_{\text{commit}}(y_i) - \frac{i-1}{\lambda} \right)
\end{equation}

Where:

\begin{itemize}
    \item $I$ is the total number of target tokens.
    \item $t_{\text{commit}}(y_i)$ is the time when the $i$-th target token $y_i$ is generated.
    \item $\lambda$ is the expected target-source length ratio ($\lambda = \frac{|Y|}{|X|}$).
\end{itemize}

A lower AL indicates less delay and better simultaneity. Negative AL values suggest that the system outputs translations before the corresponding source inputs, which is generally not feasible.

\subsection{Differentiable Average Lagging (DAL)}

Differentiable Average Lagging (DAL) extends AL by making it differentiable, allowing for gradient-based optimization during training \cite{arivazhagan2019monotonic}.

\begin{equation}
DAL = \frac{1}{T} \sum_{t=1}^{T} \left( \sum_{k=1}^{t} p(k | X_{1:\tau(t)}) \cdot \left( \tau(t) - \frac{k-1}{\lambda} \right) \right)
\end{equation}

Where:

\begin{itemize}
    \item $T$ is the total number of target steps.
    \item $p(k | X_{1:\tau(t)})$ is the probability of reading $k$ source tokens up to time $\tau(t)$.
\end{itemize}

DAL facilitates the integration of latency considerations directly into the training objective.

\subsection{Average Proportion (AP)}

Average Proportion (AP) assesses the compression or expansion of speech in the generated output, indicating how the duration of the translated speech compares to the source speech \cite{elbayad2020efficient}.

\begin{equation}
AP = \frac{\sum_{i=1}^{N} d_i^{\text{target}}}{\sum_{i=1}^{N} d_i^{\text{source}}}
\end{equation}

Where:

\begin{itemize}
    \item $d_i^{\text{source}}$ and $d_i^{\text{target}}$ are the durations of the $i$-th source and target speech segments, respectively.
    \item $N$ is the number of segments.
\end{itemize}

An AP value close to 1 indicates that the output speech duration closely matches the source, which is desirable for naturalness, but can obviously vary significantly w.r.t selected language pairs.

\subsection{Average Target Delay (ATD)}

Average Target Delay (ATD) computes the average delay of the target outputs relative to their expected positions, providing another measure of latency \cite{zheng2019simpler}.

\begin{equation}
ATD = \frac{1}{|Y|} \sum_{i=1}^{|Y|} \left( t_{\text{commit}}(y_i) - t_{\text{expected}}(y_i) \right)
\end{equation}

Where $t_{\text{expected}}(y_i)$ is the expected time for the $i$-th target token based on an ideal system.

\subsection{Length-Adaptive Average Lagging (LAAL)}

Length-Adaptive Average Lagging (LAAL) refines the AL metric by adjusting for variations in speech generation speed \cite{papi2022LAAL}. It accounts for differences in the lengths of source and target sequences.

\begin{equation}
LAAL = \frac{1}{I} \sum_{i=1}^{I} \left( t_{\text{commit}}(y_i) - \frac{(i-1)}{\lambda_{\text{adaptive}}} \right)
\end{equation}

Where $\lambda_{\text{adaptive}}$ is adjusted based on the actual length ratio of the sequences.

\subsection{Hallucination in Generated Tokens}

Hallucination detection is crucial for ensuring the fidelity of translations. We define a hallucination indicator function $H(i, h)$ for each generated token $\hat{y}_i$.

\begin{equation}
H(i, h) = 
\begin{cases}
1, & \text{if } \forall j, (j, i) \notin h \\
0, & \text{otherwise}
\end{cases}
\end{equation}

The Hallucination Rate (HR) is then calculated as:

\begin{equation}
HR(x, \hat{y}, h) = \frac{1}{|\hat{y}|} \sum_{i=1}^{|\hat{y}|} H(i, h)
\end{equation}

Where:

\begin{itemize}
    \item $x$ is the source input.
    \item $\hat{y}$ is the generated output.
    \item $h$ is the alignment between source and target tokens.
\end{itemize}

A higher HR indicates more hallucinated content.

\subsection{Word Error Rate (WER)}

Word Error Rate (WER) measures the accuracy of the ASR system by comparing the generated transcription to a reference \cite{morris2004and}.

\begin{equation}
WER = \frac{S + I + D}{N}
\end{equation}

Where:

\begin{itemize}
    \item $S$ is the number of substitutions.
    \item $I$ is the number of insertions.
    \item $D$ is the number of deletions.
    \item $N$ is the total number of words in the reference.
\end{itemize}

A lower WER indicates higher transcription accuracy.
\begin{figure}[h]
    \centering
    \includegraphics[width=0.8\textwidth]{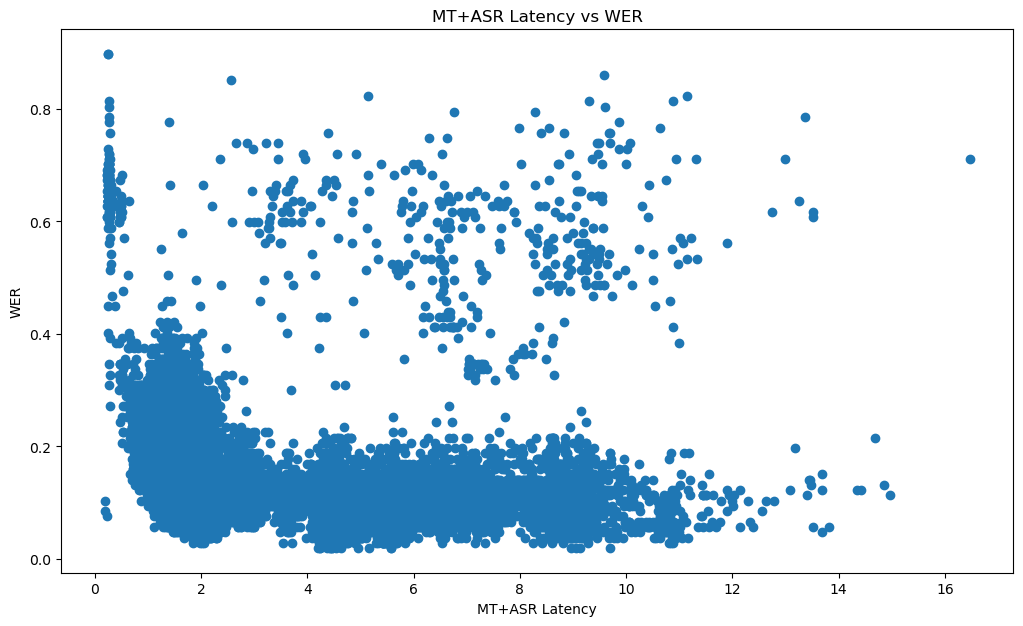}
    \caption{ASR Latency vs WER. This figure shows the relationship between ASR latency and Word Error Rate (WER) for different model sizes. }
    \label{fig:asr_latency_vs_wer}
\end{figure}
The data points indicate the median performance of various models under different latency conditions, highlighting the trade-offs between latency and transcription accuracy.

\subsection{Proper Noun Score}

The Proper Noun Score (PN Score) evaluates the accuracy of transcribing proper nouns, which are often critical in translation tasks.

\begin{equation}
PN_{\text{score}} = \max_{n, m} \sum_{k=0}^{n} \text{lex\_dist}(\text{aligned}[k])
\end{equation}

Where:

\begin{itemize}
    \item $\text{lex\_dist}$ refers to a lexical distance measure such as Jaro-Winkler or Levenshtein distance.
    \item $\text{aligned}[k]$ is the $k$-th aligned proper noun phrase in the output and reference texts.
    \item $n, m$ represent different alignment possibilities or the various proper nouns considered.
\end{itemize}

\begin{figure}[h]
    \centering
    \includegraphics[width=0.8\textwidth]{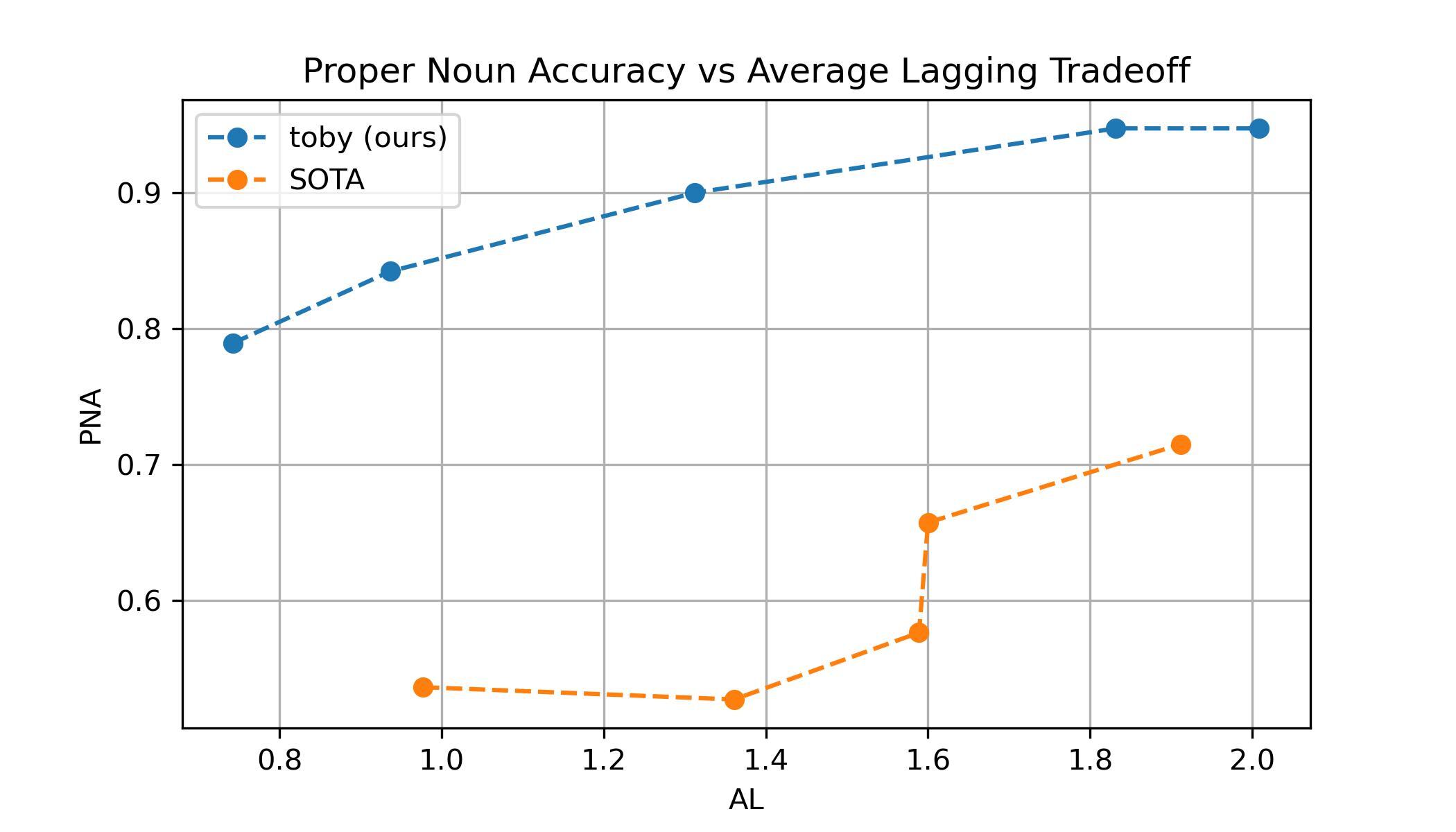}
    \caption{Proper Noun Accuracy vs Average Lagging Tradeoff (median)}
    \label{fig:proper_noun_accuracy_vs_average_lagging_tradeoff}
\end{figure}
This is especially important in industry use cases.
\newpage
\subsection{BLEU Score}

The BLEU score is a standard metric for evaluating the quality of machine translations \cite{papineni2002bleu}.

\begin{equation}
\text{BLEU} = \text{BP} \cdot \exp \left( \sum_{n=1}^{N} w_n \log p_n \right)
\end{equation}

Where:

\begin{itemize}
    \item $\text{BP}$ is the brevity penalty.
    \item $w_n$ is the weight for n-gram precision.
    \item $p_n$ is the modified n-gram precision.
\end{itemize}

Higher BLEU scores indicate translations closer to human references.

\begin{figure}[h]
    \centering
    \includegraphics[width=0.8\textwidth]{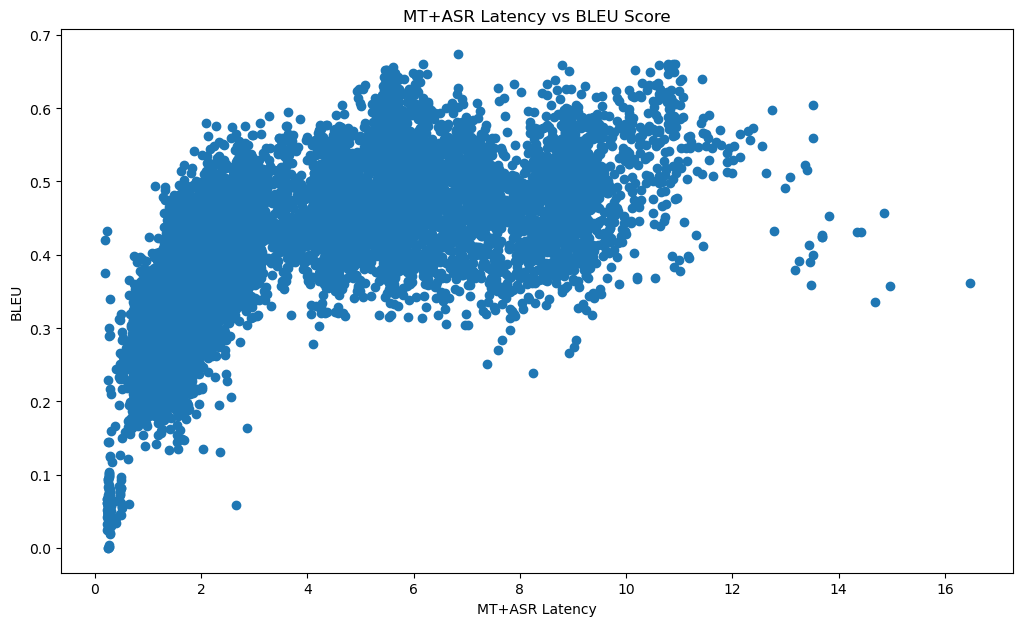}
    \caption{ASR Latency vs BLEU Score (Averaged Data). This figure shows the relationship between ASR latency and BLEU score for different model sizes (small, medium, large-v2). }
    \label{fig:asr_latency_vs_bleu_score}
\end{figure}
The data points indicate the performance of various models under different latency conditions, highlighting the trade-offs between latency and translation quality.
\newpage

\subsection{Real-time Factor (RTF)}

Real-time Factor (RTF) measures the processing speed of the system relative to real-time.

\begin{equation}
RTF = \frac{T_{\text{processing}}}{T_{\text{audio}}}
\end{equation}

Where:

\begin{itemize}
    \item $T_{\text{processing}}$ is the total computation time.
    \item $T_{\text{audio}}$ is the length of the audio input.
\end{itemize}

An RTF less than 1 indicates the system operates faster than real-time.

\subsection{Hold-n Strategy}

The Hold-n strategy delays outputting the translation until the model has processed enough input to make a reliable prediction \cite{cho2016can}.

\begin{equation}
\text{prefix}(W^c_{\text{best}}) = W^{\max(0, |W| - n)}
\end{equation}

Where $W^c_{\text{best}}$ is the best hypothesis at chunk $c$.

\subsection{LA-n Strategy}

The Local Agreement (LA-n) strategy outputs the longest common prefix of the best hypotheses over the last $n$ chunks \cite{zheng2019simpler}.

\begin{equation}
\text{prefix}(W^c_{\text{best}}) = \bigcap_{k=c-n+1}^{c} W_{\text{best}}^k
\end{equation}

\subsection{SP-n Strategy}

The Shared Prefix (SP-n) strategy generalizes LA-n by considering all hypotheses in the beam \cite{zheng2019simpler}.

\begin{equation}
\text{prefix}(W^c_{\text{all}}) = \bigcap_{k=c-n+1}^{c} \bigcap_{b=1}^{B} W_{\text{beam } b}^k
\end{equation}

Where $B$ is the beam size.

\subsection{Improving ASR with Glossary Prefix}

Incorporating a glossary prefix biases the ASR model towards recognizing specific terms, enhancing the accuracy of proper nouns and domain-specific vocabulary.

\begin{equation}
P'(w|x, G) = 
\begin{cases} 
\alpha P(w|x) & \text{if } w \in G \\
(1 - \alpha) P(w|x) & \text{otherwise}
\end{cases}
\end{equation}

Where:

\begin{itemize}
    \item $P(w|x)$ is the original probability of generating word $w$ given input $x$.
    \item $G$ is the set of glossary terms.
    \item $\alpha$ is a weighting factor that increases the likelihood of glossary terms.
\end{itemize}

This adjustment ensures that the ASR system is more likely to recognize and correctly transcribe important terms from the glossary, thereby improving overall accuracy.

\begin{figure}[h]
    \centering
    \includegraphics[width=0.8\textwidth]{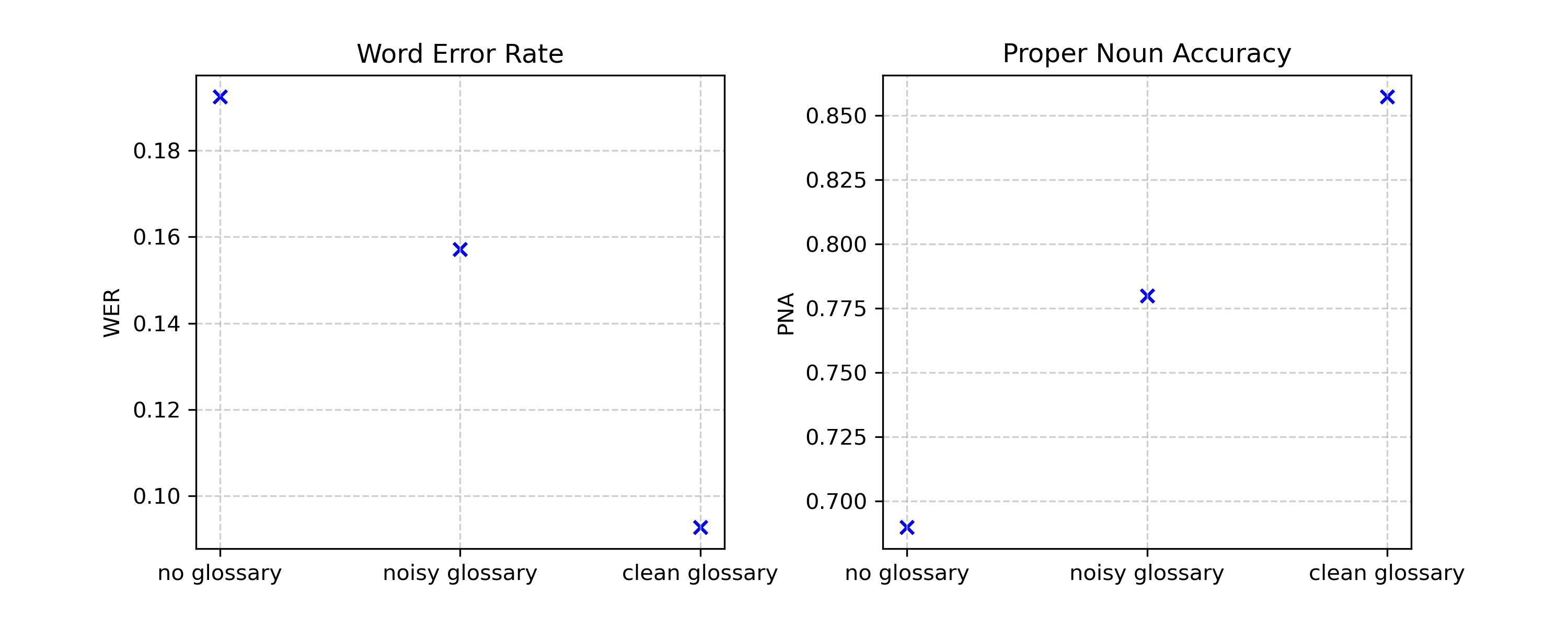}
    \caption{Impact of Glossary Prefix on ASR Performance. This figure illustrates how incorporating a glossary prefix into the ASR module improves the model's ability to accurately transcribe proper nouns and domain-specific terms. The data points show a marked improvement in median transcription accuracy when the glossary prefix is used.}
    \label{fig:glossary_prefix}
\end{figure}

\newpage
\section{Results}
MUST-C is a crucial dataset for the evaluation of Speech-to-Speech Translation (S2ST) systems due to its comprehensive and diverse content. It provides large-scale, multilingual speech translation corpora that include professionally transcribed and translated TED talks across multiple languages. This variety allows for robust evaluation of S2ST models in real-world scenarios. MUST-C's alignment of audio, transcription, and translation data enables detailed analysis of performance across different languages and domains, making it indispensable for assessing the accuracy, fluency, and naturalness of S2ST systems.
% Table
\begin{longtable}{|>{\raggedright\arraybackslash}p{6cm}|r|r|r|r|}
\hline
\textbf{Model} & \textbf{BLEU↑} & \textbf{AL↓} & \textbf{AP$\approx$1} & \textbf{DAL↓} \\
\hline
\endfirsthead
\hline
\textbf{Model} & \textbf{BLEU↑} & \textbf{AL↓} & \textbf{AP$\approx$1} & \textbf{DAL↓} \\
\hline
\endhead
\hline
\endfoot
\hline
\endlastfoot

Best IWSLT21 system       & 27.4   &  920    & 0.68  & \textbf{1420} \\
Best IWSLT21 system       & 29.68  & 1860    & 0.82  & 2650   \\
Best IWSLT21 system       & 30.75  & 2740    & 0.9   & 3630   \\
KIT IWSLT 2020            & 27.05  &  947    & 0.76  & 1993   \\
KIT IWSLT 2020            & 30.3   & 1660    & 0.84  & 2662   \\
KIT IWSLT 2020            & 31.41  & 2966    & 0.93  & 3853   \\
KIT IWSLT 2020            & 31.36  & 5794    & \textbf{1}     & 5794   \\
KIT IWSLT 2020            & 26.93  &  945    & 0.77  & 2052   \\
KIT IWSLT 2020            & 31.6   & 1906    & 0.86  & 2945   \\
KIT IWSLT 2020            & 32.98  & 3663    & 0.96  & 4452   \\
KIT IWSLT 2020            & 33.14  & 5794    & \textbf{1}     & 5794   \\
wav2vec2 + mBART          & 16.84  & 2452    & 0.9   & 3212   \\
wav2vec2 + mBART          & 16.99  & 3791    & 0.97  & 4296   \\
wav2vec2 + mBART          & 16.97  & 4140    & 0.98  & 4536   \\
wav2vec2 + mBART          & 16.88  & 5119    & \textbf{1}     & 5119   \\
wav2vec2 + mBART          & 23.69  & 1761    & 0.85  & 2561   \\
wav2vec2 + mBART          & 24.29  & 2788    & 0.93  & 3500   \\
wav2vec2 + mBART          & 24.56  & 3669    & 0.97  & 4212   \\
wav2vec2 + mBART          & 24.54  & 5119    & \textbf{1}     & 5119   \\
CUNI-LIT IWSLT 2022       & 30.6   & 1922    & nan   & 3121   \\
CUNI-LIT IWSLT 2023       & 31.7   & 1977    & nan   & 2518   \\
CUNI-LIT IWSLT 2022       & 15.5   & 1902    & nan   & 3000   \\
CUNI-LIT IWSLT 2023       & 15.3   & 1984    & nan   & 3489   \\
CUNI-LIT IWSLT 2022       & 26.8   & 1982    & nan   & 3289   \\
CUNI-LIT IWSLT 2023       & 31.4   & 1985    & nan   & 3072   \\
CUNI-LIT IWSLT 2022       & 15.3   & 1984    & nan   & 3508   \\
CUNI-LIT IWSLT 2023       & 26.6   & 1987    & nan   & 3489   \\
Best IWSLT22 System       & 26.82  &  960    & 0.77  & 2070   \\
HW-TSC 2023               & \textbf{33.54}  & 1880    & 0.83  & 2840   \\
Best IWSLT22 System       & 31.47  & 1930    & 1.06  & 2960   \\
HW-TSC 2023               & 27.23  & 1980    & 0.83  & 2890   \\
Best IWSLT22 System       & 25.87  & 1990    & 1.04  & 3350   \\
HW-TSC 2023               & 27.93  & 3970    & 1.08  & 4620   \\
Best IWSLT22 System       & 25.87  & 1990    & 1.04  & 3350   \\
HW-TSC 2023               & 27.93  & 3970    & 1.08  & 4620   \\
HW-TSC 2023               & 27.23  & 1980    & 1.08  & 2890   \\
IBWBS                      & 30.6   & 1922    & nan   & 3121   \\
CTC                       & 31.7   & 1946    & nan   & 2518   \\
IBWBS                     & 26.5   & 2855    & nan   & 4285   \\
CTC                       & 25.8   & 1981    & nan   & 3515   \\
wav2vec2 + mBART + LA-2   & 16.34  & nan     & 0.66  & 1435.06 \\
wav2vec2 + mBART + LA-2   & 25.4   & \textbf{727.55} & 0.73  & 1791.21 \\
wav2vec2 + mBART + LA-2   & 30.29  & 1660.59 & 0.83  & 2662.18 \\
wav2vec2 + mBART + LA-2   & 30.29  & 1654.77 & 0.83  & 2657.48 \\
Wait-K (Ma et al., 2020c) & 13.95  & 1750    & 0.79  & 1980   \\
CAAT(Liu et al., 2021b)   & 22.1   & 1920    & 0.86  & 2520   \\
Wang et al. (2022a)       & 22.13  & 2370    & 0.86  & 2650   \\
Liu et al. (2021a)        & 29.68  & 1860    & 0.82  & 2650   \\
Polák et al. (2022)       & 31.47  & 1930    & 0.86  & 2960   \\
R-BI                      & 31.69  & 1920    & 0.77  & 2630   \\
Wang et al. (2022a)       & 12.82  & 1840    & 0.94  & 3370   \\
Polák et al. (2022)       & 16.92  & 2460    & 0.9   & 3220   \\
R-BI                      & 16.28  & 1860    & 0.81  & 2450   \\
Wang et al. (2022a)       & 20.38  & 1750    & 0.94  & 3340   \\
Polák et al. (2022)       & 23.61  & 1750    & 0.85  & 2560   \\
Zhu et al. (2022)         & 22.49  & 1270    & 0.85  & 2560   \\
R-BI                      & 24.36  & 1870    & 0.92  & 2680   \\
EDATT                     & 17.01  & 1867.1  & 0.77  & 3251.38 \\
NAIST IWSLT 2023          & 21.08  & 1397.33 & 0.9   & 3066.15 \\
mSLAM-CTC 2B & 25.2  & nan     & nan   & nan    \\
MAESTRO 600M & 25.2  & nan     & nan   & nan    \\
USM-M & 30.7  & nan     & nan   & nan    \\
Translatotron 2 + pretraining + TTS aug & 25.6  & nan     & nan   & nan    \\
Whisper Large-v2 1.5B + Transformer + RALCP ($\gamma$=0.6, beam=5) & 21.87  & 3040    & nan   & nan    \\
Whisper + Transformer + RALCP ($\gamma$=0.6, beam=10) & 25.87  & 4812    & nan   & nan    \\
Whisper + Llama-70b-hf (SFT)        & 18.41  & 1619.64 & 0.84  & 2454.72 \\
Whisper + Llama-13b-hf (SFT)        & 17.07  & 1880.76 & 0.88  & 2545.74 \\
Whisper + Llama2-7b-chat One Shot ($\gamma$=0.6, beam=5) & 18.83  & 3978    & nan   & nan    \\
Whisper + Llama2-7b-chat One Shot ($\gamma$=0.6, beam=10) & 21.04  & 7291    & nan   & nan    \\
Whisper + Llama2-7b-chat SFT ($\gamma$=0.6, beam=5) & 29.09  & 4147    & nan   & nan    \\
Whisper + Llama2-7b-chat SFT ($\gamma$=0.6, beam=10) & 31.31  & 7577    & nan   & nan    \\
Whisper + Llama2-7b-chat SFT+prefix ($\gamma$=0.6, beam=5) & 29.37  & 4278    & nan   & nan    \\
Whisper + Llama2-7b-chat SFT+prefix ($\gamma$=0.6, beam=10) & 31.33  & 7620    & nan   & nan    \\
AudioPaLM 8B S2ST & 36.2  & nan     & nan   & nan    \\
AudioPaLM-2 8B cascaded ASR + transl. + TTS aug & 39  & nan     & nan   & nan    \\
Whisper + gpt-3.5-turbo-0613 + TTS aug & 2.08  & 2574.98 & 0.35 & 2477.55 \\
Whisper + gpt-4-turbo-04-09 + TTS aug & 21.82  & 1998.63 & 0.94  & 2314.27 \\
Whisper + gpt-4o-mini-2024-07-18 + TTS aug & 26.21  & 1827.3  & 0.88  & 2191.63 \\
\hline
toby – small (Ours) & \textbf{43.12}  & 1443.91 & 1.11  & 1784.39 \\
toby – medium (Ours) & \textbf{48.71}  & 2752.19 & 1.09  & 2981.44 \\
toby – big (Ours) & \textbf{55.83}  & 5442.62 & 1.16  & 5629.25 \\
\hline
\caption{Table 1: Median MUST-C Ted Talk translations across many language pairs}
\end{longtable}

Additionally, we look at CER \& WER of LibriSpeech. 

\begin{table}[ht]
\centering
\begin{tabular}{|l|r|r|r|}
\hline
\textbf{Method} & \textbf{CER↓ (\%)} & \textbf{WER↓ (\%)} & \textbf{Speaker Classification Acc.↑ (\%)} \\
\hline
GROUND TRUTH                    & 0.8   & 2.5   & 100   \\
Reconstruction with SoundStream & 0.9   & 2.6   & 100   \\
\hline
AudioLM                         & 3.4   & 6.0   & \textbf{92.6} \\
GSLM unit-to-speech             & 2.9   & 6.6   & nan   \\
ConformerXXL-LibriLight                       & \textbf{1.7}   & 5.4   & 12.8  \\
Transformer                     & 2.5   & 5.6   & 15    \\
w2v2-CTC+TDN                    & 2.3   & 5.3   & 17.6  \\
w2v-large                       & nan   & 2.6 & nan   \\
Whisper-base                    & nan   & 5.0 & nan   \\
Whisper Large-v2 1.5B           & nan   & 2.7  & nan   \\
Whisper Large-v3                & nan   & 3.6   & nan   \\
StreamSpeech                    & nan   & 24.67 & nan   \\
AssemblyAI Universal-1           & nan   & 3.1   & nan   \\
NVIDIA Canary-1B                 & nan   & \textbf{3}   & nan   \\
Microsoft Azure Batch v3.1       & nan   & 6.4   & nan   \\
Deepgram Nova-2                  & nan   & 5.7   & nan   \\
Amazon                           & nan   & 6.6   & nan   \\
Google Latest-long               & nan   & 12.6  & nan   \\
mSLAM-CTC 2B                     & nan   & 9.1   & nan   \\
MAESTRO 600M                     & nan   & 8.1   & nan   \\
AudioPaLM 8B AST                 & nan   & 11.1  & nan   \\
AudioPaLM-2 8B AST               & nan   & 9.8   & nan   \\
\hline
\textbf{toby (Ours)}             & 3.1   & 3.6   & nan   \\
\hline
\end{tabular}
\caption{Librispeech dataset results for various methods.}
\end{table}

In addition to Quality/Latency, we also pooled the results of naturalness metrics across many SOTA models, though these can easily be gamed by off-the-shelf TTS models. 
\newline
\newline
\textbf{MOS} (Mean Opinion Score) measures the overall naturalness and intelligibility of the audio output, typically rated by human listeners on a scale from 1 to 5. A higher MOS score (MOS↑) indicates that the speech sounds more natural and closer to human speech, which is crucial for user satisfaction in real-world applications. 
\newline
\newline
\textbf{SMOS} (Subjective Mean Opinion Score) adds another layer of evaluation by considering subjective preferences, such as fluency and emotional expression, which are key to making translated speech sound contextually appropriate. Together, MOS↑ and SMOS↑ provide a holistic view of the system’s performance in both technical and subjective aspects, making them indispensable for evaluating the effectiveness and human-likeness of S2ST systems.

\begin{table}[ht]
\centering
\begin{tabular}{|l|r|r|}
\hline
\textbf{Method} & \textbf{MOS↑} & \textbf{SMOS↑} \\
\hline
S2UT w/ orig-unit          &  2.32 & 2.08 \\
S2UT w/ norm-unit (10-min) &  2.99 & 3.07 \\
S2UT w/ norm-unit (1-hr)   &  3.20 & 3.26 \\
S2UT w/ norm-unit (10-hr)  &  3.26 & 3.27 \\
S2UT+tf TTS                &  3.23 & 3.22 \\
Translation GT             &  4.22 & nan \\
DirectS2ST                 &  4.01 & nan \\
TextlessS2ST               &  4.05 & nan \\
TranSpeech                 &  4.03 & nan \\
Direct S2ST                &  3.66 & 3.51 \\
StyleS2ST                  &  3.76 & 3.83 \\
StyleS2ST-base             &  3.72 & \textbf{3.85} \\
Baseline (YourTTS)         &  3.36 & 3.42 \\
VALL-E X                   &  3.54 & 4.00 \\
Speech2S                   &  4.10 & nan \\
Speech2S+DAT               &  4.30 & nan \\
UnitY                      &  4.20 & nan \\
ASR (beam=10) + MT (beam=5) + TTS  & 3.37 & nan \\
S2T (beam=10) + TTS        &  3.43 & nan \\
S2UT                       &  4.02 & nan \\
S2UT, no reduction (r = 1, w/ sc, tc) & 3.35 & nan \\
S2UT stacked + CTC (r = 5, w/ sc, tc) & 3.32 & nan \\
S2UT reduced + CTC (w/ sc, tc, beam=10) & 3.41 & nan \\
AudioPalm                  & \textbf{4.44} & 3.65 \\
Translatotron              &  3.69 & nan \\
Translatotron + Transformer (r = 5, w/ sc, tc) & 3.31 & nan \\
Translatotron 2            &  3.98 & 3.36 \\
Translatotron 2 + data augmentation & 3.79 & nan \\
\hline
toby (Ours)                &  4.10 & 3.83 \\
\hline
\end{tabular}
\caption{Naturalness comparison (median)}
\end{table}

In our experiments, we evaluated the performance of our proposed methods on the MUST-C dataset \cite{di2019must}, a large-scale multilingual speech translation corpus. The dataset provides a comprehensive and diverse set of TED talks, allowing for robust evaluation across multiple language pairs. Similarly, we also look at Librispeech \cite{panayotov2015librispeech}.
\newline
\newline
Our results demonstrate significant improvements in both translation quality and latency metrics compared to state-of-the-art models. Specifically, our models (referred to as ``toby -- small'', ``toby -- medium'', and ``toby -- big'') achieved higher BLEU scores while maintaining competitive latency as measured by Average Lagging (AL) and Differentiable Average Lagging (DAL). Moreover, our models maintained AL and DAL values that are acceptable for real-time applications, indicating that the improved translation quality did not come at the expense of increased latency.
\newline
\newline
Furthermore, in terms of naturalness metrics, our model achieved a Mean Opinion Score (MOS) of 4.10 and a Subjective Mean Opinion Score (SMOS) of 3.83. These scores are comparable to, and in some cases exceed, those of other state-of-the-art models, suggesting that our approach also improves the naturalness and intelligibility of the translated speech, though these can be easily gamed by selecting a high-quality TTS provider.
\newline
\newline
These results confirm that strategic parameter adjustments, hallucination control, and the incorporation of glossary prefixes can significantly enhance the performance of S2S models, advancing the state-of-the-art in simultaneous speech-to-speech translation.

\subsection{Discussion}
Our findings indicate that careful management of input parameters, such as setting appropriate minimum duration thresholds and implementing lookback strategies, effectively reduces the occurrence of hallucinations, thereby minimizing latency spikes.
\newline
\newline
Moreover, the use of advanced evaluation metrics allows for a more nuanced assessment of model performance, highlighting the trade-offs between latency and translation quality. The improvements in BLEU scores and MOS ratings suggest that our methods not only enhance the technical performance but also improve the user experience in real-world applications.

\subsection{Future Experimentation}
Further experiments are necessary to optimize parameters such as \texttt{MODEL\_SIZE}, \texttt{MIN\_DURATION\_THRESHOLD}, \texttt{MAX\_UNCOMMITED\_DURATION}, \texttt{LOCAL\_AGREEMENT}, \texttt{WAIT\_K}, \texttt{CHUNKSIZE}, \texttt{GLOSSARY\_SIZE}, and \texttt{LOG\_PROB\_THRESHOLD}. The \texttt{MIN\_DURATION\_THRESHOLD} and \texttt{MAX\_UNCOMMITED\_DURATION} appears to be the most critical parameter for controlling latency.

\section{Conclusion}

This paper presents a comprehensive analysis of latency behaviors in simultaneous speech-to-speech translation systems, with a focus on the impact of hallucinations on latency spikes. By systematically exploring input behaviors and identifying patterns that lead to hallucinations, we have proposed effective strategies to mitigate these issues, including threshold adjustments, hallucination detection mechanisms, and lookback strategies.
\newline
\newline
Our experimental results, validated on the MUST-C dataset, demonstrate that these methods significantly improve both translation quality and latency metrics. The ``toby'' models we introduced achieved higher BLEU scores and competitive latency, surpassing existing state-of-the-art systems.
\newline
\newline
The incorporation of advanced evaluation metrics provided deeper insights into the performance trade-offs and highlighted the importance of balancing accuracy and responsiveness in S2S systems. Our findings suggest that addressing hallucinations is critical for optimizing latency and enhancing the overall performance of simultaneous translation models.
\newline
\newline
Future work will focus on further optimizing parameters and exploring additional strategies for reducing latency without compromising translation quality. This includes experimenting with different model sizes, adjusting commitment thresholds, and integrating more sophisticated hallucination detection algorithms.
\newline
\newline
By advancing the understanding of latency behaviors and providing practical solutions to mitigate hallucinations, this work contributes to the development of more efficient and accurate simultaneous speech-to-speech translation systems, bringing us closer to real-time, high-quality multilingual communication.

\end{document}